\documentclass[ conference]{IEEEtran}

\IEEEoverridecommandlockouts 

\title{\LARGE \bf
Collaborative Multi-agent Stochastic Linear Bandits
}

\author{Ahmadreza Moradipari$^{1}$,  Mohammad Ghavamzadeh$^{2}$ and Mahnoosh Alizadeh$^{1}$
\thanks{This work is supported by NSF grant 1847096.}
\thanks{$^{1}$University of California, Santa Barbara,
       {\tt\small ahmadreza$\_$moradipari, alizadeh@ucsb.edu} }%
\thanks{$^{2}$Google Research
{\tt\small ghavamza@google.com}}%
        
}

\usepackage[ruled,vlined,linesnumbered,noresetcount ]{algorithm2e}
\usepackage{verbatim}
\usepackage{amsmath,amssymb,amsfonts,amsthm}
\usepackage{mathtools}  
\usepackage{tabulary}
\usepackage{booktabs}
\usepackage{graphicx}
\usepackage{graphics}
\usepackage{ textcomp }
\usepackage{pdfpages}
\usepackage{colortbl}
\usepackage[english]{babel}
\usepackage{lipsum}
\usepackage{ nicefrac}
\usepackage[noend]{algpseudocode}
 \usepackage{bbm} 
\usepackage{xcolor}

\usepackage{dsfont}

\usepackage[hidelinks]{hyperref}
\definecolor{darkred}{RGB}{150,0,0}
\definecolor{darkgreen}{RGB}{0,150,0}
\definecolor{darkblue}{RGB}{0,0,150}
\hypersetup{colorlinks=true, linkcolor=red, citecolor=blue, urlcolor=darkblue}

\newtheorem{assumption}{Assumption}
\newtheorem{theorem}{Theorem}[section]

\newtheorem{lemma}[theorem]{Lemma}

\newtheorem{remark}{Remark}[section]

\newcommand{\norm}[1]{\left\lVert#1\right\rVert}
\usepackage{color}

\begin{document}

\maketitle

\begin{abstract}
    We study a collaborative multi-agent stochastic linear bandit setting, where $N$ agents that form a network communicate locally to minimize their overall regret.  In this setting, each agent has its own linear bandit problem (its own reward parameter) and the goal is to select the best global action w.r.t. the average of their reward parameters. At each round, each agent proposes an action, and one action is randomly selected and played as the network action. All the agents observe the corresponding rewards of the played action, and use an accelerated consensus procedure to compute an estimate of the average of the rewards obtained by all the agents. We propose a distributed upper confidence bound (UCB) algorithm and prove a high probability bound on its $T$-round regret in which we include a linear growth of regret associated with each communication round.  Our regret bound is of order $\mathcal{O}\Big(\sqrt{\frac{T}{N \log(1/|\lambda_2|)}}\cdot (\log T)^2\Big)$, where $\lambda_2$ is the second largest (in absolute value) eigenvalue of the communication matrix.
\end{abstract}

\section{Introduction}

Stochastic linear bandits have been extensively studied in decision-making problem with a linear structure, e.g., recommendation systems or path routing \cite{bubeck2012regret,moradipari2021parameter,moradipari2018learning}. At each time step, an agent plays an action and receives its corresponding random reward with an expected value that is linearly dependent on the action's context. The goal of the agent is to collect as much reward as possible over $T$ rounds.

In this work, we consider the stochastic linear bandit problem in a multi-agent setting, where a team of $N$ agents cooperate locally on a network to maximize their collective reward. Specifically, we study  the setting where each agent has its own linear bandits problem, i.e., the underlying reward parameter varies among the agents, and the network's \textit{true} reward parameter is averaged among all the agents. Therefore, agents need to collaborate  in order to maximize the global welfare of the network. Moreover, we want to avoid excessive communication among the agents, and hence we assume that 1) agents can only share their information with their neighbors in the network; 2) each communication step translates into a linear growth of regret. For this case, we prove a regret of order $\mathcal{O}(\sqrt{\frac{T}{N \log(1/|\lambda_2|)}})$ which depends on the spectral gap  of the structure matrix (i.e., $1-|\lambda_2|$) and decreases with the size of the network $N$.

As we review next, multi-agent bandit problems have received significant attention in the past few years.
{ As a motivational example, consider a fashion brand that is connected to $N$ influencers over different locations. Here, influencers represent the agents in our setting. The goal of the company is to maximize its global welfare, i.e., to agree upon the best product (i.e., action) with respect to the whole network’s true reward parameter with the help of its influencers.  The effect of different products on their respective customer base is unknown a-priori to the influencers. Therefore,   each influencer has  their  own  linear  bandit  problem  (its own  reward  parameter), which represents the preference of the customers with whom they are interacting. In this decentralized setting, each   influencer only receives feedback from their own customers, but other influencers in their network can observe their score of the recommended  product (reward signal of the played action), e.g., see how many likes they have received for a post. Hence, agents collaborate with their neighbors in order to collectively move towards the best action with respect to the network's true parameter (i.e., the best product that fits the whole network). This decentralized setting has also been motivated by \cite{shahrampour2017multi} for the multi-armed bandit framework, with applications in selecting a brand from a finite list of brands for a social network, where each agent based on its own preferences votes for a brand (i.e., action) in the given list of finite number of brands, and the brand with the majority of votes has been selected with the goal of maximizing social welfare. This model is also applicable in distributed detection systems where agents need to identify the global unknown parameter for which they only have partial information \cite{shahrampour2015distributed}.}

The contributions of this work are as follows: we study a variant of the decentralized multi-agent linear bandits for a (connected) network structure of $N$ agents who can only communicate with their immediate neighbors. The network's true reward parameter is averaged across all agents and as such cannot be learnt separately by individual agents. For this, we propose and analyze a fully-decentralized UCB-based algorithm, and adopt the accelerated consensus procedure for communication between the agents so that each agent can compute an estimate of the network's true reward parameter. We then prove that even with a limited number of communication rounds, agents are able to construct high probability confidence regions that include the network's reward parameter. We show that this comes with an extra cost in the regret,
and we discuss the trade-off between regret and communication. Moreover, we theoretically and empirically show the effect of size of the network as well as the spectral gap of the network's graph on regret.  

\subsection{Related work}\label{lit:review}

\textbf{Multi-armed Bandits (MAB).}
Two popular algorithms exist for MAB: 1)    the   upper confidence bound (UCB) algorithm based on the optimism in the face of uncertainty (OFU) principle \cite{Auer}, which chooses the best feasible environment and corresponding optimal action at each time step with respect to  confidence regions on the unknown parameter; 2) Thompson Sampling (TS) (a.k.a., posterior sampling) algorithm~\cite{thompson1933likelihood}, which samples an environment from the prior at each time step and selects the optimal action with respect to the sampled parameter. 

\textbf{Linear Bandits.}
In stochastic linear bandit (LB) problems, actions are modeled by feature vectors and the expected reward corresponding to each action is a linear function of its feature vector. Two well-known algorithms for LB are: linear UCB (LinUCB) and linear Thompson Sampling (LinTS). \cite{abbasi2011improved} provided a regret bound of order $\mathcal{O}(\sqrt{T} \log T)$ for LinUCB, and \cite{abeille2017linear}, \cite{moradipari2020stage}, \cite{moradipari2021safe}, and \cite{moradipari2020linear} provided a regret bound of order $\mathcal{O}(\sqrt{T} (\log T)^{3/2})$ for LinTS in a frequentist setting, where the unknown reward parameter $\theta_\star$ is fixed. 

\textbf{Multi-agent Stochastic Bandits.}
There has been  increasing attention on studying   distributed or decentralized  bandit problems in the past years. 
In the multi-armed bandits settings, the work of \cite{landgren2016distributed} study the problem of distributed multi-armed bandits. They propose two UCB-based algorithms, called coop-UCB and coop-UCB2. In coop-UCB, it is assumed that  all the eigenvalues and corresponding eigenvectors of the structure matrix are known, which is a stronger assumption. In \cite{9143736}, they consider a multi-agent multi-armed bandits setting, where they also consider a cost for communication between agents and propose an efficient sampling rule and a communication protocol for each agent to maximize its own total expected cumulative reward. In \cite{chakraborty2017coordinated}, each agent can either send her information to the whole network or pull an arm. There has been another line of work that consider the distributed MAB problem with collision, i.e., if two or more agents  choose the same arm at the same time, they either receive no reward or the reward splits among them \cite{LANDGREN2021109445,anandkumar2011distributed}.
\cite{kar2011bandit} considers a setting where at each time step only one agent can play an action and observes the corresponding reward, while the others have only access to the information that has been sent by that agent.
In \cite{shahrampour2017multi}, whose setting is the closest to ours, the authors consider a MAB problem where each agent has a different reward distribution over each arm, and the total regret is minimized with respect to the global best arm, i.e., the one with the largest expected reward when averaged across the  agents. In their setting, the arm  played at each round is chosen by the majority vote of the agents, which restricts their setting to the finite action set. We generalize their setting to LB, and hence, we can handle both finite and infinite action sets. Another limitation of their setting is that the actions and their corresponding rewards are independent across players and arms (because in MAB the rewards are independent between arms). However, in our setting, we model actions by their feature vectors, and thus, they could be dependent on each other in some sense. For example, in music recommendation systems, two different genre  of music, such as rock and metal are not completely independent. The recent work of \cite{martinez2019decentralized} studies the setting where the agents are solving a similar MAB problem and they can only share their information with their neighbors. They use the Chebyshev acceleration applied to the consensus procedure \cite{arioli2014chebyshev} for communication between agents, so that each agent can compute an approximation of the mean reward of each arm over all the network only by sharing information with their neighbors. In our paper, we adopt similar communication protocols in Section~\ref{sec:alg_decrribtion}. In \cite{korda2016distributed}, the authors study a setting in which at each round, each agent can only share its information with a randomly selected agent that is not necessarily among its neighbors. \cite{gagrani2018thompson} study the two agent team-learning problem under two problem settings of decoupled dynamics with no information sharing and  coupled dynamics with delayed information sharing.

 \textbf{Notation.}  The weighted $\ell_2$-norm with respect to a positive semi-definite matrix $V$ is denoted by $\|x\|_V = \sqrt{x^{\top} V x}$. 
Let $\mathcal{F}_t = (\mathcal{F}_1, \sigma ( x_1,\xi_1, \dots, x_t,\xi_t))$  be the filtration ($\sigma$-algebra) that represents the information up to round $t$. Eigenvalues of the structure matrix $W$ are noted by $1 = \lambda_1 > |\lambda_2| \geq \dots \geq |\lambda_N|$. They are sorted by their absolute value. 


\section{Problem Setting}\label{sec:setting:multiagent}
We consider  a multi-agent network of $N$ agents that sequentially select actions played by the whole network. The network is represented by an undirected  graph $G$, and  agents are represented by the set of nodes $\mathcal{V} = \{ 1,\dots,N\}$ on the graph. We assume that the graph $G$ is connected, i.e., there exists a path from each agent $i$ to any other agent $j$.  In our setting, each agent can only communicate with its neighbors. We define the doubly stochastic graph structure matrix $W$ with non-negative entries (i.e., $W_{i,j} \geq 0, \forall i,j$) in order to capture the interaction between the agents with respect to graph $G$. Therefore, $W_{ij}=0,$ if there is no edge in $G$ that connects node $j$ to node $i$. Each agent has its own local linear bandit problem (with different reward parameter), and communicates information with its neighbors on the graph in order to solve the global bandit problem. Next, we first describe the bandit problem that agents are solving, and then explain the network structure that we adopt.

\textbf{Local and Global Bandit Problems.} Each agent is given a set of actions $\mathcal{D}$. At each round $t$, by playing the action $x_t$,  each agent $i$ observes a random reward \begin{align}
    r_t^i = \langle x_t, \theta_\star^i \rangle + \xi_t^i, ~\forall i = 1,\dots,N \label{reward:each_agent_i},
\end{align} 
where $\theta_\star^i$ is the \textit{unknown}, but fixed, reward parameter and $\xi_t^i$ is an additive zero-mean random noise of agent $i$. Note that for agents $i$ and $j$, the vectors $\theta_\star^i$ and $\theta_\star^j$ are not equal in general. We further assume that the \textit{true} global reward parameter of the network is defined as: 
\begin{align}
    \mu_\star = \frac{1}{N} \sum_{i = 1}^N \theta_\star^i. \label{true_reward_parameter_of_network}
\end{align}
Agents aim to select the action that maximizes the \textit{global} expected reward of the network (i.e., with respect to $\mu_\star$). Since each agent $i$  has only access to the observation based on her own reward parameter $\theta_\star^i$, it cannot make good decisions in order to maximize the global expected reward of the network without collaborating with the other agents. Therefore, agents must communicate with each other in order to estimate the true reward parameter $\mu_\star$ in \eqref{true_reward_parameter_of_network}. In our setting, we assume that at each round $t$, the network coordinator randomly selects an agent index $a(t) \in \mathcal{V}$ and all the agents in the network play the action proposed by the agent $a(t)$, i.e.,  an action $x_t^{a(t)} \in \mathcal{D}$ is played by all the agents. Then, each agent $i$ observes a reward $r_t^i$ according to \eqref{reward:each_agent_i}.     
We define the \textit{cumulative pseudo-regret} of the network as \begin{align}
    R(T) := \sum_{t=1}^T \langle x_\star, \mu_\star \rangle - \langle x_t^{a(t)}, \mu_\star \rangle, \label{def:regret_netowrk}
\end{align}
where $x_\star$ is the optimal action that maximizes the global expected reward of the network, i.e., $x_\star = \arg\max_{x \in \mathcal{D}} \langle x, \mu_\star \rangle$. Note that maximizing the global expected reward is equivalent to minimizing the pseudo-regret of the network. Henceforward, we use regret to refer to the pseudo-regret $R(T)$. { We note that in our decentralized setting, the network coordinator can only propagate the selected action (e.g., a product) to the $N$ agents (e.g., influencers), but it does not receive any feedback from the social network and it does not perform any computation since the number of samples could be very large and the computation could be very costly.}

\textbf{Network Structure.} We assume that the structure matrix $W$ is symmetric and  the sum of each row, and hence the sum of each column is $1$. This implies that 1 is an eigenvalue of $W$. We further assume that all the eigenvalues of $W$ are real and less than one in absolute value, i.e., $1 = \lambda_1 > |\lambda_2| \geq \dots \geq |\lambda_N| \geq 0$. For our algorithm, similar to \cite{martinez2019decentralized}, we only allow agents to have limited information about the graph's structure. Specifically, we assume that each agent can only share its information with its neighbors, and that it knows the total number of nodes (i.e., agents) as well as the absolute value of the second largest eigenvalue of matrix $W$ i.e., $|\lambda_2|$. This is certainly less information compared to the existing work in multi-agent LB, as described in Section \ref{lit:review}. 

{ Our algorithm relies on decentralized communication between neighboring agents in the network so as to allow them to compute estimates of the average of rewards obtained by all the agents for the action taken by the netwrk $x_t^{a(t)}$. Details of how agents propose the network action $x_t^{a(t)}$ and the communication protocol used to calculate the average reward is provided next. First, we state some standard assumptions. }
\begin{assumption}\label{ass:sub-gaussian_noise}
For all $t$ and $i = 1,\dots,N$, $\xi_t^i$ is conditionally zero-mean R-sub-Gaussian noise variables, i.e., $\mathbb{E}[\xi_t^i | \mathcal{F}_{t-1}] = 0$, and $\mathbb{E}[e^{\lambda \xi_t^i} | \mathcal{F}_{t-1}] \leq \exp{(\frac{\lambda^2 R^2}{2})}, \forall \lambda \in \mathbb{R}$, and { $\mathcal{F}_t = (\mathcal{F}_1, \sigma ( x_1,\xi_1, \dots, x_t,\xi_t))$ is the the history of the entire network up to round $t$.}
\end{assumption}
\begin{assumption}\label{ass:bounded_parameter}
There exists a positive constant $S$ and $L$ such that  $\|\theta_{\star}^i \|_2 \leq S$, and hence $\|\mu_{\star} \|_2 \leq S$, and $\|x\|_2 \leq L, \forall x \in \mathcal{D}$. Also, we assume $\langle x, \mu_{\star} \rangle \in [-1,1], \forall x \in \mathcal{D}$. 
\end{assumption}


\section{Algorithm Description}\label{sec:alg_decrribtion}

To tackle our multi-agent linear bandit problem, we propose a modified version of the linear UCB-based algorithm proposed in \cite{abbasi2011improved} that we refer to as MA-LinUCB. The pseudo-code of the MA-LinUCB is presented in Algorithm \ref{alg:multi-agent_LB}. MA-LinUCB is an episodic algorithm such that each episode consists of two phases: 1) an exploration-exploitation phase, 2) a communication phase. MA-LinUCB plays actions only during the exploration-exploitation phase, and during the communication phase, agents only share information with their neighbors. As we will explain later, this approach allows us to quantify the cost of unnecessarily lengthy communication, as a lack of action translates into a linear growth of regret due to no reward being accrued.  

\textbf{Exploration-exploitation Phase.}
Let $t_s$ be the time when the episode $s$ begins. At time $t_s$ the network coordinator randomly selects an agent index $a(s)$. Then,  agent $a(s)$, given a regularized least-square estimate $\hat{\mu}_{t_s}^{a(s)}$ for the global network parameter using the information up to time $t_s$,  constructs the ellipsoidal confidence region $\mathcal{E}_{t_s}^{a(s)}$, and   computes the best feasible action-environment pair ($\tilde{x}_{t_s}^{a(s)}, \tilde{\mu}_{t_s}^{a(s)}$) according to the LinUCB action selection rule (line 6). The optimistic action $\tilde{x}_{t_s}^{a(s)}$ is chosen as the network action  $x_{t_s} = \tilde{x}_{t_s}^{a(s)}$ and all the agents play that action. By playing the network action, each agent observes its corresponding reward according to \eqref{reward:each_agent_i}. Then MA-LinUCB activates the communication phase.

\textbf{Communication Phase.} {During the communication phase, agents do not play any action and they only share information with their neighbors in order to collectively find {\it an approximation} of the average reward all agents in the network have received during episode $s$, i.e., $\frac{1}{N}\sum_{i=1}^N r_{t_s}^i$. This means that for the length of the communication phase, agents do not receive any reward, and as such regret grows linearly with the length of the communication phase,  highlighting a trade-off between how accurately the agents can estimate the average network reward,  and the regret of a longer communication phase. The communication protocol allows each agent $i$ to share her latest  reward  $r_{t_s}^i$ with her neighbors according to the structure matrix $W$, receive their reward signals, and update her information according to an accelerated consensus procedure in \cite{martinez2019decentralized}.  The summary of how the agents share their information in the accelerated consensus procedure, i.e., the $\mbox{mix}$ function in line 12 of Algorithm 1,  is presented in Algorithm 2 of \cite{martinez2019decentralized}, and we omit it here due to limited space. }  When the communication phase of episode $s$ finishes, each agent $j$ can compute $y_s^j$,  {\it an approximation} of the average rewards over all the agents in the network at episode $s$, i.e., $\frac{1}{N}\sum_{i=1}^N r_{t_s}^i$. Then, each agent updates her RLS estimate using this  approximation. The accuracy of the approximation depends on the length of the communication phase $q(s)$, which we carefully optimize to control regret.

Lemma \ref{polynomial_of_matrix} (from \cite{martinez2019decentralized}) stated next captures how accurate the approximation of the average observed reward will be after certain communication steps following an accelerated consensus procedure. 

\begin{lemma}\label{polynomial_of_matrix}
Let $W$ be a structure matrix with real eigenvalues such that $ \mathds{1}^\top W = \mathds{1}^\top$ and $W \mathds{1} = \mathds{1}$ and all the eigenvalues are less than one in absolute value.   Fix $\epsilon > 0$, and let $q(\epsilon) = \lceil \log(2N/\epsilon) / \sqrt{2 \log(1/|\lambda_2|)} \rceil$ where $\lambda_2$ is the second largest eigenvalue in absolute value of matrix $W$. Then after $q(\epsilon)$ communication time steps based on Algorithm 2 of \cite{martinez2019decentralized}, each agent can construct a polynomial $p_{q(\epsilon)}(W)$ of the structure matrix $W$ that satisfies $\| p_{q(\epsilon)}(W) - \frac{1}{N}\mathds{1} \mathds{1}^\top \|_2 \leq \epsilon/N $.
\end{lemma}

{
Moreover, let $r_s = [r_{t_s}^i]_{i=1,\dots,N}$ be the $N$-dimensional vector containing the observed reward of each agent at episode $s$ before any communication.  Then,  after $q(\epsilon) = \lceil \log(2N/\epsilon) / \sqrt{2 \log(1/|\lambda_2|)} \rceil$ communication steps, agents deploying Algorithm 2 of \cite{martinez2019decentralized} can compute a polynomial $p_{q(\epsilon)}(W)$ of the structure matrix $W$ applied to the  vector $r_s$, i.e., $p_{q(\epsilon)}(W) r_s$. Here, the $j$-th entry of the vector, i.e., $(p_{q(\epsilon)}(W)r_s)_j$, is the  approximation $y_s^j$ available to agent $j$ for the average of the values in reward vector $r_s$, i.e., $\frac{1}{N} \sum_{i=1}^N r_{t_s}^i$. 
 For a fixed $\epsilon$, we can use Lemma \ref{polynomial_of_matrix} to write the accuracy of this approximation as $
     \| p_{q(\epsilon)}(W) r_s - \frac{1}{N}\mathds{1}  \mathds{1}^\top r_s  \|_2 \leq  \| p_{q(\epsilon)}(W) - \frac{1}{N}\mathds{1} \mathds{1}^\top \|_2 \| r_s \|_2   \leq  \frac{\epsilon}{ N} N = \epsilon$, where according to Assumption \ref{ass:bounded_parameter}, we have $\|r_s\|_2 \leq N$.}
Note that each entry $j$ of the vector $p_{q(\epsilon)}(W) r_s $ is computed by agent $j$ using only hers and her neighbors' reward information. The only other information needed to compute the Chebyshev polynomials $p_{q(\epsilon)}(W)$ (see details in \cite{scaman2017optimal}) are the total number of agents $N$  and second largest eigenvalue in absolute value of the structure matrix $W$, i.e., $\lambda_2$. 

\begin{algorithm} 
\caption{Multi-agent Linear UCB (MA-LinUCB)}\label{alg:multi-agent_LB}
\textbf{Input:}  $\delta, T, \lambda, t = 1, s = 1$;

\While{$t\leq T$ }{
Set $q(s) = \lceil \log(2N s) / \sqrt{2 \log(1/|\lambda_2|)} \rceil$;

start time of episode $s: t_s \leftarrow t$ \\

The network coordinator selects the agent $a(s)$\\

Agent $a(s)$ computes:
$(\tilde{x}_{t_s}^{a(s)}, \tilde{\mu}_{t_s}^{a(s)}) = \arg\max_{x \in \mathcal{D}}\max_{\theta \in \mathcal{E}_{t_s}^{a(s)}} \langle x, \theta \rangle$

Agents play the network action $x_{t_s}= \tilde{x}_{t_s}^{a(s)}$\\

\textbf{for each agent} $i$: observe $r_{t_s}^i = \langle x_{t_s}, \theta_\star^i\rangle + \xi_{t_s}^i$ \\

\textbf{if} $t_s + q(s) > T$ \textbf{then} return

\textbf{else} activate communication phase:

\For{  $h= 0,\dots, q(s)-1$}{
$r_{t_s}^i \leftarrow$ mix($r_{t_s}^i$,$h$,$i$)\\
}\textbf{end for}

$y_{s}^i \leftarrow r_{t_s}^i$

\textbf{end if} 

$t \leftarrow t_s + q(s)$

Update the RLS-estimates $\hat{\mu}_{t}^i$ according to \eqref{RLS_estimates}\\
Build the confidence region: $\mathcal{E}_t^i (\delta) = \{ v \in \mathbb{R}^d : \norm{ v-\hat{\mu}_{t}^i}_{V_{t}} \leq \beta_t(\delta)   \}$
\\
$s \leftarrow s+1$
}\textbf{end while}
\SetAlgoLined
\end{algorithm}

Clearly, a longer communication phase allows agents to compute a more accurate approximation  of the average rewards of the network but also adversely affects regret, presenting a design trade-off. A requirement of any UCB-based algorithm like ours is that the agents should be able to construct high probability ellipsoidal confidence regions  $\mathcal{E}^i_t$ that include the global reward parameter $\mu_\star$ at all rounds.  An important observation is that this requirement cannot be met through a communication phase with a constant length. We need to carefully design the   length of the communication phase to just be long enough so that each agent   can construct a high probability confidence region that includes $\mu_\star$, but not any longer so that we do not adversely affect the network regret. 
As such, at the heart of Algorithm \ref{alg:multi-agent_LB} and its proof of regret lies an analytic argument that materialized the intuition described above. 
The length of the communication phase for episode $s$ of the algorithm is denoted by $q(s)$. Hence, for episode $s+1$, we have $t_{s+1} = t_s + q(s)$. We show that by choosing the length of the communication phase of the episode $s$ to be $q(s) = 1 + \log(2N s) / \sqrt{2 \log(1/|\lambda_2|)}$, we can guarantee the high probability confidence region as well as control the regret of the network.
Note that the length of the communication phases is increasing in time ($q(t) \propto \log t$).
Intuitively, in the first rounds, agents do not have  accurate estimates of their underlying reward parameters, and hence their current information is not very useful to share,  which leads to the short length of the communication phases. As the algorithm progresses, since each agent can  compute a better approximation of her underlying reward parameter, and hence they can share useful information, the length of communication phases is longer. { Moreover, we can conclude that the number of times that communication happens is at most of order $\mathcal{O}(\frac{T \sqrt{2\log(1/|\lambda_2|)}}{\log(2N)})$.}

After $q(s)$ communication steps, agents update their regularized least-square (RLS) estimates at time $t = t_s + q(s)$ as follows: \begin{align}
\hat{\mu}_t^i = V_{t}^{-1} \sum_{k=1}^{s} x_{t_k} y_k^i, ~\text{where}~ V_{t} = \lambda I + \sum_{k =1}^{s} x_{t_k} x_{t_k}^{\top}, \label{RLS_estimates}
\end{align} where  $x_{t_k}$ is the network action  at episode $k$, and $y_k^i$ is the estimation of the reward available to the agent $i$ after communication phase of episode $k$. We note that MA-LinUCB only updates the RLS-estimates when the communication phase finishes, and agents have computed an approximation of the average of the rewards. We will see in Theorem \ref{confidence_region} that this is critical for constructing the ellipsoid confidence regions that contain the global reward parameter $\mu_\star$.

\section{Regret Analysis}\label{sec:analyisofregret}

We first recall that each agent $i$'s underlying reward parameter is $\theta_\star^i$. However, the goal of the agents is to maximize the network's global reward  whose underlying parameter is averaged over all the agents, i.e., $\mu_\star$. It turns out that  to handle this issue, each agent needs to compute the average of rewards of all the agents in the network. However, due to communication limitations, each agent can only receive the reward signals of her neighbors in the network. The key idea that we use is to add a controlled number of communication phases to the algorithm so agents can share their information. In particular, after observing their reward signal, agents share their information using an accelerated consensus procedure. Then, each agent can compute an approximation of the average network reward.
Next, in Theorem \ref{confidence_region}, we show that by using the approximation of the average of the rewards, each agent  can build a modified version of the confidence ellipsoid given in \cite{abbasi2011improved} such that { it includes the network's true parameter $\mu_\star$ with high probability.} 

\begin{theorem}\label{confidence_region}
Let Assumptions \ref{ass:sub-gaussian_noise}, \ref{ass:bounded_parameter} hold. Fix any $\delta \in (0,1)$, and let the structure matrix $W$ satisfy the conditions required in Lemma \ref{polynomial_of_matrix}. Furthermore, let $t_s$ be the start time of episode $s$. We choose the length of the communication  phase as $q(s) = \lceil \log(2N s) / \sqrt{2 \log(1/|\lambda_2|)} \rceil$. Then, at each time $t$ in the interval $[t_s,t_{s+1})$ and using the information up to round $t_s$, each agent $i$ can construct a confidence region $\mathcal{E}_t^i$  that includes the parameter $\mu_\star$ with probability at least $1-\delta$ as: 
\begin{align}
  \mathcal{E}_t^i (\delta) = \{ v \in \mathbb{R}^d : \norm{ v-\hat{\mu}_{t}^i}_{V_{t}} \leq \beta_t(\delta)   \}, \label{confidenceradiusbound}
\end{align} where $\beta_{t}(\delta) =  \frac{R}{\sqrt{N}} \sqrt{d \log\left( \frac{1 + s L^2 /\lambda }{\delta} \right) } + \sqrt{\lambda } S + L/\sqrt{\lambda}$.
 \end{theorem}
 
Therefore, we show that agents only require an approximation of the average rewards to construct the confidence regions that include the true global reward parameter $\mu_\star$ with high probability. However, we show that, this comes with an extra cost, i.e., $L/\sqrt{\lambda}$ in \eqref{confidenceradiusbound}.  Now we are ready to bound the overall regret of the network. 
In Theorem \ref{thm:regretofalgorithm}, we provide a regret bound for MA-LinUCB. 

\begin{theorem}\label{thm:regretofalgorithm}
Let Assumptions \ref{ass:sub-gaussian_noise}, \ref{ass:bounded_parameter} hold,   $\lambda \geq \max(1,L^2)$, and the structure matrix $W$ satisfy the condition mentioned in Lemma \ref{polynomial_of_matrix}. For a fixed $\delta \in (0,1)$, the regret defined in \eqref{def:regret_netowrk} for Algorithm \ref{alg:multi-agent_LB} is upper bounded with probability at least $1-\delta$ by \begin{align}
R(T) \leq & 4 \beta_T(\delta) \sqrt{2 T' d \log{(1 \!+\! \frac{T' L^2}{d \lambda})}} (1  \!+\! \frac {\log(2N T)} { \sqrt{2 \log(1/|\lambda_2|)}}) \nonumber\\& +  \log(2N T) / \sqrt{2 \log(1/|\lambda_2|)}
\end{align} where $T' = \frac{T}{1+q(1)} $, $q(1)= \frac{\log(2N)}{\sqrt{2\log(1/|\lambda_2|)}}$.
\end{theorem}
 
\begin{remark}
The upper bound provided in Theorem \ref{thm:regretofalgorithm} on the T-period regret for Algorithm \ref{alg:multi-agent_LB}  has the following order: \begin{align}
    R(T) \leq \mathcal{O} \left(  d  \frac{\log(NT)}{\sqrt{N}} \log\left(\frac{TL^2}{\delta}\right) \sqrt{\frac{T}{\log(1/|\lambda_2|)}} \right). \nonumber
\end{align}
\end{remark}

\begin{remark}
The regret bound provided in Theorem \ref{thm:regretofalgorithm} depends on the spectral gap of the structure matrix. Recall the spectral gap of the structure matrix $W$ as $SG(W) = 1 - |\lambda_2|$. Therefore, for  networks with larger spectral gap, our algorithm performs better.
\end{remark}

Intuitively, the overall regret of the network can be decomposed into two terms: regret caused by the exploration-exploitation phases, and that of communication phases.  We bound each term separately. The idea of bounding term I is to use the result of Theorem \ref{confidence_region}, and bound the regret for the times that algorithm plays the optimistic actions. We first find an upper bound on the number of times the network plays optimistic actions, and then bound the regret of it. For the second term, we know that during the communication phase, agents only share information and they do not play any action, and hence regret grows linearly. However, in light of our results on the length of the communication phases,  we bound the regret of the second term with logarithmic order. In particular, from the Assumption \ref{ass:bounded_parameter}, we have $\langle x_\star, \mu_\star \rangle \leq 1$. Therefore, the regret of the communication phase of episode $s$ of the algorithm is at most $q(s)$. Hence, we  can bound the regret of the second term by the maximum length of the communication phases up to time $T$, i.e., $q(T)$. 

\section{Numerical Results}\label{sec:numerical_results}
In this section, we investigate the numerical performance of our proposed algorithm on synthetic data. We show the effect of the size and spectral gap of the network on the performance of our algorithms. In all implementations, we used $\delta = 1/4 T$, $r=0.1$, and $\mathcal{D} = [-1,1]^4$. The reward parameters are drawn from $\mathcal{N}(0,I_4)$. We have implemented a modified version of LinUCB which uses $\ell_1$-norms instead of $\ell_2$-norms due to computational considerations (for details see \cite{Dani08stochasticlinear}).
The structure matrix $W$ is chosen according to \cite{duchi2011dual}. 
We first define the graph Laplacian matrix as $\mathcal{W} = I - M^{1/2} A M^{-1/2}$, where $A$ is the adjacency matrix of the graph $G$, and $M$ is a diagonal matrix such that $M_{ii}$ is the degree of node $i$. For non-regular graphs, if we call the maximum degree of the nodes $k_{\max}$, we choose $W =  I - \frac{1}{k_{\max} + 1 }  M^{1/2} \mathcal{W} M^{1/2}$. For $k$-regular graphs, we choose $W = I - \frac{k}{1+k} \mathcal{W}$. 

\begin{figure}[h]
     \centering
          \includegraphics[width=0.35\textwidth]{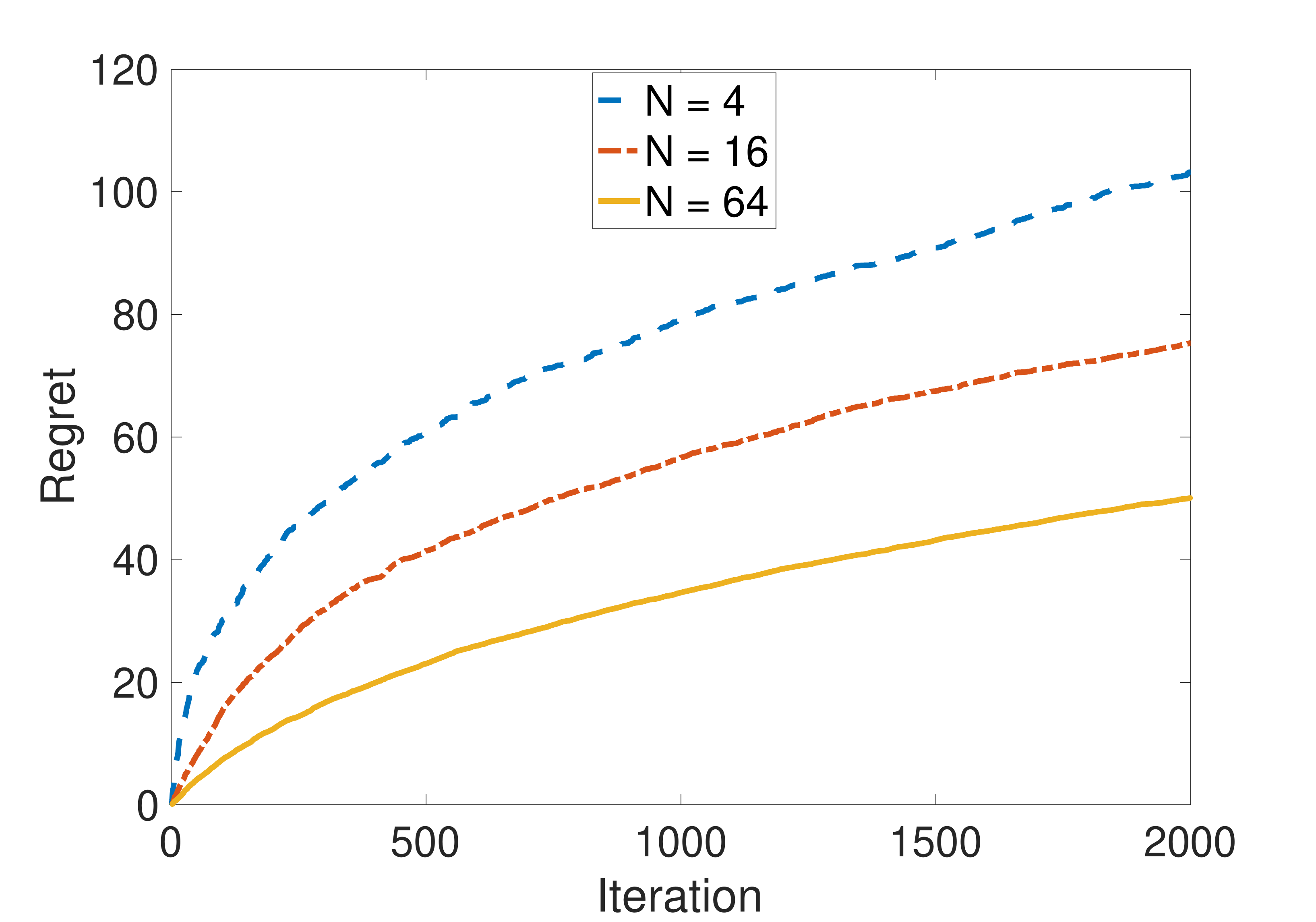}
         \caption{Comparison of the average cumulative regret of of  MA-LinUCB for a complete graph with different number of agents.}
         \label{fig:differentnumbers}
  \end{figure}

In Figure \ref{fig:differentnumbers},  we plot the average cumulative regret of   MA-LinUCB on a complete graph (with self loops) for different network sizes $N=4,16,64$ over 100 realizations. We show in Theorem \ref{thm:regretofalgorithm} that regret decreases with the order $\mathcal{O}(\frac{\log N}{\sqrt{N}})$ in the number of agents in network. Similar to \cite{shahrampour2017multi}, we plot Figure \ref{fig:differentnumbers}  in complete graphs in order to remove the effect of the spectral gap of the matrix (the spectral gap of the complete network is $1$).


\begin{figure}[h]
     \centering
          \includegraphics[width=0.35\textwidth]{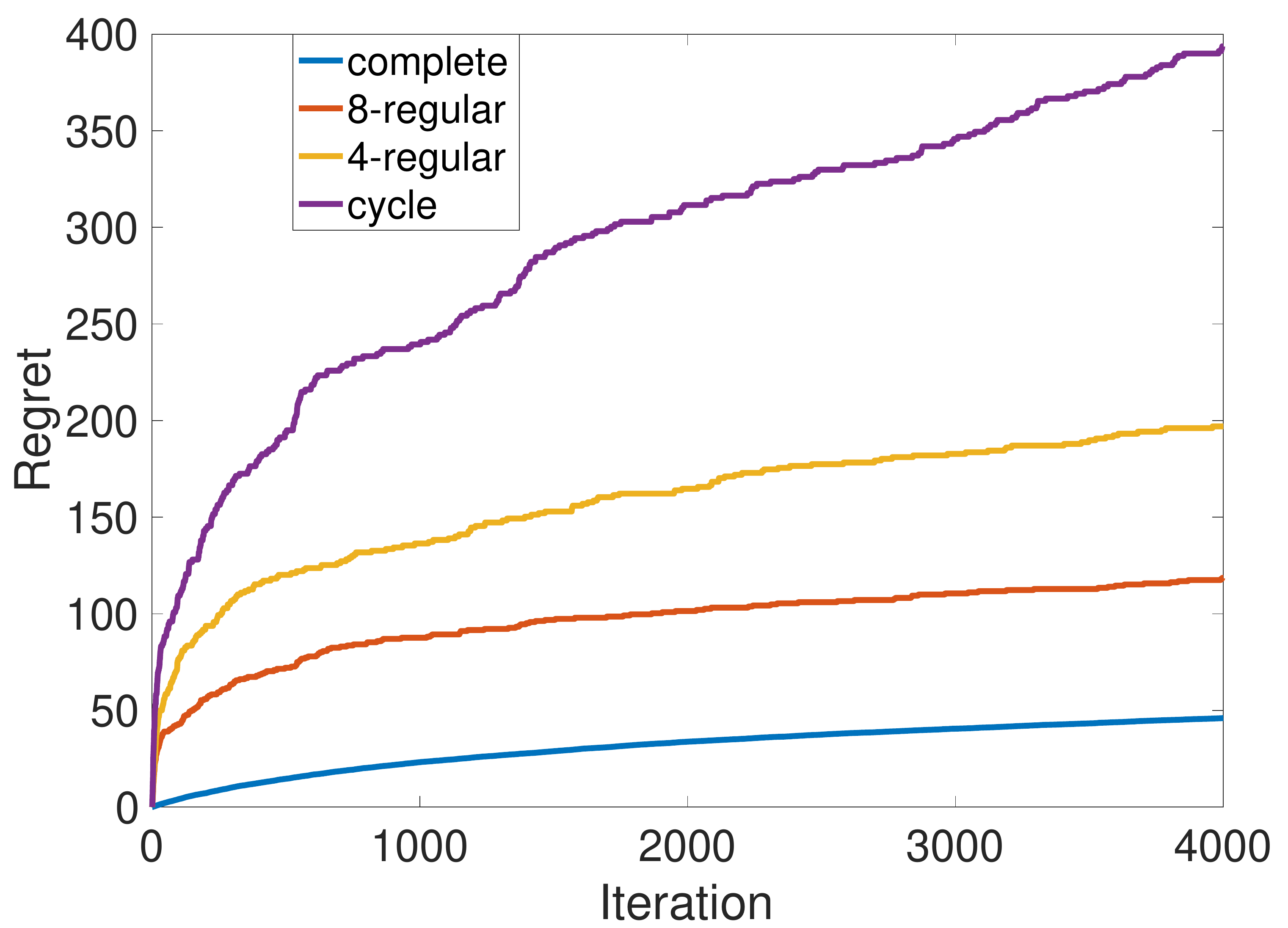}
         \caption{Comparison of the average cumulative regret of MA-LinUCB for complete and $4$-regular and $8$-regular and cycle graphs. }
         \label{fig:differentgraph}
  \end{figure}

In Figure \ref{fig:differentgraph}, we fix the number of agents in the network to $N=50$. We evaluate the performance of the MA-LinUCB in different graphs: complete, $8$-regular, $4$-regular, and cycle (all with self-loops). The values in Figure \ref{fig:differentgraph} are averaged over 100 realizations. Figure \ref{fig:differentgraph} captures the effect of the spectral gap of the structure matrix on the cumulative regret. Recall the definition of the spectral gap as $SG(W) = 1 - |\lambda_2|$. As it is shown, networks with the larger spectral gap have smaller regret. In particular, for complete network with the largest spectral gap has the smallest regret in comparison to the $k$-regular networks. Also, the cycle network has smaller spectral gap than $k$-regular networks for $k >2$.

 \vspace{-0.25cm}
\section{Conclusion}
In this work, we studied a variant of the collaborative multi-agent stochastic linear bandits problem. In particular, each agent has its own linear bandit problem (its own reward parameter) and the goal is to select the best global action w.r.t. the average of their reward parameters. Therefore, agents need to collaborate to estimate the global reward parameter. At each round,   one action is randomly selected and proposes  the so-called network action. We adopt the accelerated consensus procedure as a communication scheme between agents and carefully design its length to control network regret. We proposed a distributed UCB-based algorithm with provable regret guarantee. For future work, a natural extension of our setting is  time-varying graphs or generalized linear bandits. 
{  Additionally, it is interesting to study extensions of setting to recently studied ``incentive-compatible bandits'' in \cite{mansour2020bayesian} where, in order to maximize the social welfare, the network coordinator should incentivize the agents to follow its policy, otherwise,
they can act selfishly in order to maximize their own reward.}

 \vspace{-0.35cm}

\bibliographystyle{IEEEtran}
\bibliography{bibfile}

\end{document}